\documentclass[conference]{IEEEtran}
\IEEEoverridecommandlockouts
\usepackage{cite}
\usepackage{amsmath,amssymb,amsfonts}
\usepackage{algorithmic}
\usepackage{graphicx}
\usepackage{textcomp}
\usepackage{xcolor}
\usepackage{caption}
\usepackage{algorithm}
\usepackage{algorithmic}
\usepackage{graphicx}
\usepackage{subcaption}
\usepackage{url}

\usepackage{tikz}
\usetikzlibrary{shapes.geometric, arrows, positioning}

\tikzstyle{block} = [rectangle, draw, fill=blue!20, 
                     text width=6em, text centered, rounded corners, minimum height=3em]
\tikzstyle{arrow} = [thick,->,>=stealth]

\usepackage{booktabs}

\def\BibTeX{{\rm B\kern-.05em{\sc i\kern-.025em b}\kern-.08em
    T\kern-.1667em\lower.7ex\hbox{E}\kern-.125emX}}
\begin{document}

\title{ A Tiered GAN Approach for Monet-Style Image Generation
\\}

\author{\IEEEauthorblockN{FNU Neha}
\IEEEauthorblockA{\textit{Dept. of Computer Science} \\
\textit{Kent State University}\\
Kent, OH, USA\\
neha@kent.edu}
\and
\IEEEauthorblockN{ Deepshikha Bhati}
\IEEEauthorblockA{\textit{Dept. of Computer Science} \\
\textit{Kent State University}\\
Kent, OH, USA \\
dbhati@kent.edu}
\and
\IEEEauthorblockN{ Deepak Kumar Shukla}
\IEEEauthorblockA{\textit{Rutgers Business School} \\
\textit{Rutgers University}\\
Newark, New Jersey, USA  \\
ds1640@scarletmail.rutgers.edu}
\and
\IEEEauthorblockN{ Md Amiruzzaman}
\IEEEauthorblockA{\textit{Dept. of Computer Science} \\
\textit{West Chester University}\\
West Chester, PA, USA \\
mamiruzzaman@wcupa.edu}
}

\maketitle

\begin{abstract}
Generative Adversarial Networks (GANs) have proven to be a powerful tool in generating artistic images, capable of mimicking the styles of renowned painters, such as Claude Monet. This paper introduces a tiered GAN model to progressively refine image quality through a multi-stage process, enhancing the generated images at each step. The model transforms random noise into detailed artistic representations, addressing common challenges such as instability in training, mode collapse, and output quality. This approach combines downsampling and convolutional techniques, enabling the generation of high-quality Monet-style artwork while optimizing computational efficiency. Experimental results demonstrate the architecture’s ability to produce foundational artistic structures, though further refinements are necessary for achieving higher levels of realism and fidelity to Monet’s style. Future work focuses on improving training methodologies and model complexity to bridge the gap between generated and true artistic images. Additionally, the limitations of traditional GANs in artistic generation are analyzed, and strategies to overcome these shortcomings are proposed.

\end{abstract}


\section{Introduction}
\begin{figure}[ht]
    \centering
    \includegraphics[width=1.0\columnwidth]{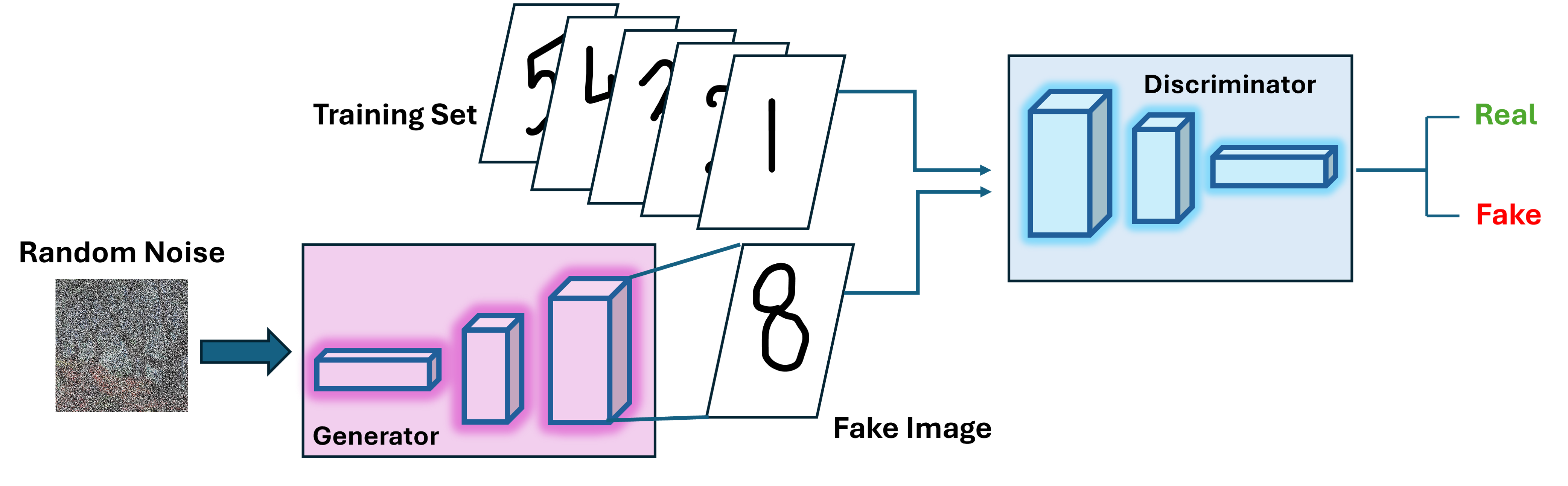}\vspace{-5pt}
    \caption{An overview of how GAN model works}\vspace{-15pt}
    \label{fig:general_gan} 
\end{figure}

Generative Adversarial Networks (GANs) have gained significant attention in machine learning, particularly in applications such as image generation, style transfer, and data augmentation. Introduced by Ian Goodfellow in 2014, GANs consist of two neural networks—a generator ($G$) and a discriminator ($D$)—engaged in a zero-sum game \cite{goodfellow2020generative,hu2024tackling}. The generator ($G$) creates synthetic data, while the discriminator ($D$) evaluates its authenticity relative to a given dataset, such as Monet paintings \cite{chang2024enhancing}. This adversarial process drives both networks to iteratively improve, enabling $G$ to produce increasingly indistinguishable outputs (see Figure \ref{fig:general_gan}). As a result, GANs have significantly advanced the generation of realistic images, videos, and even text \cite{cai2021generative,sharma2024generative}. In medical imaging, GANs enhance diagnostic image resolution and generate synthetic data to supplement limited datasets \cite{bhati2024survey}. However, the complexity of deep learning models poses challenges in understanding their decision-making processes, necessitating interpretability and visualization techniques to improve transparency \cite{bhati2024survey}.

\begin{figure}[ht]
    \centering
    \includegraphics[width=1.0\columnwidth]{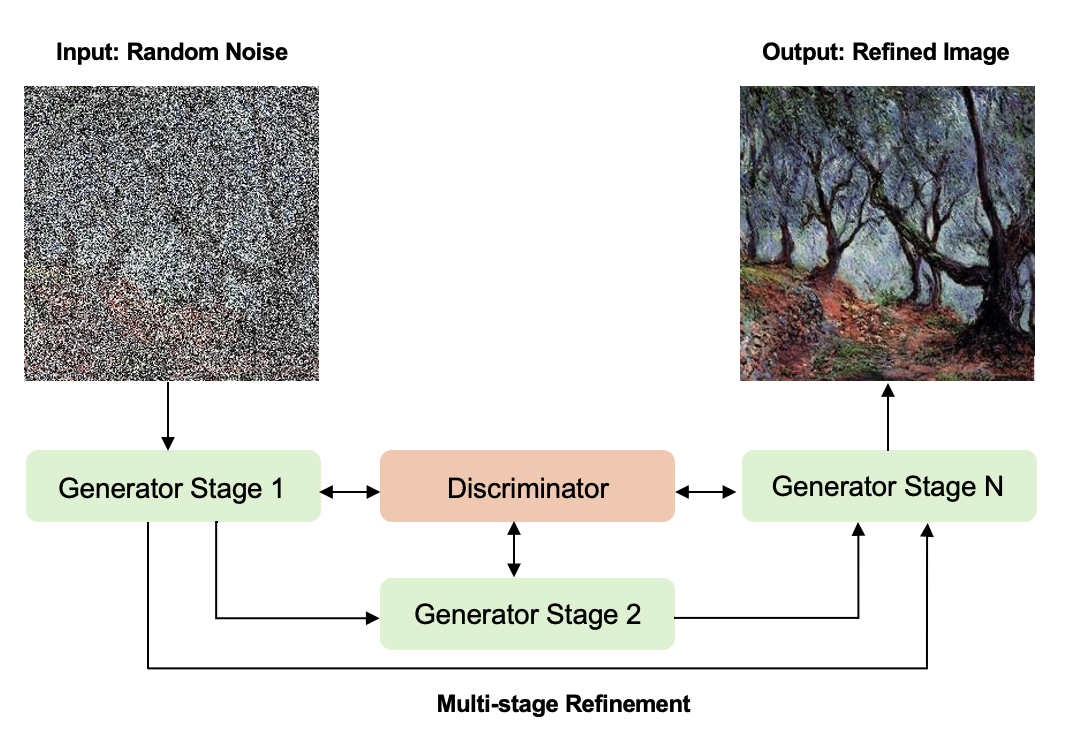}\vspace{-5pt}
    \caption{A tiered GAN model: Transforming random input to refined output. The left image displays the initial random noise, and the right image shows the refined image. This highlights the model's ability to enhance and transform images effectively.}\vspace{-15pt}
    \label{fig:refined} 
\end{figure}

Building on the success of GANs in generating realistic and artistic imagery, this work leverages their capabilities to create images that mimic the style of Claude Monet \cite{shi2024relu}. Inspired by the kaggle competition \textit{I’m Something of a Painter Myself} \cite{jang2020painter}, the proposed approach utilizes a tiered GAN model to progressively refine image quality. Starting from random noise, the model generates images through multiple stages, with each stage producing outputs that are increasingly detailed and stylistically accurate, as shown in Figure \ref{fig:refined}. This multi-stage process ensures that the $G$ learns to replicate Monet's distinctive style, enabling the creation of high-quality artistic images \cite{vela2023improving,jin2022retracted}. 

The key contributions of this work are as follows:
\begin{itemize}
    \item The proposed \textit{tiered GAN} model employs multiple GANs in succession to progressively enhance image quality, transforming low-quality inputs into refined representations of Monet's style.
    \item An \textit{efficient training method} is introduced, incorporating downsampling and convolutional layers to enable high-quality artistic generation while optimizing computational resources.
    \item An \textit{experimental analysis of GAN limitations} in artistic style generation is provided, highlighting the shortcomings of traditional GAN architectures and proposing potential improvements.
\end{itemize}

The paper is organized as follows: Section 2 presents the background. Section 3 describes GAN architecture. Section 4 discusses the methodology. Section 5 presents the results and discussion. Section 6 concludes this work with future directions.

\section{Background}
The field of artistic image generation has progressed significantly with the advent of GANs, which excel at creating visually compelling images. These models are extensively used in tasks such as style transfer and image refinement. This section explores key advancements in GAN architectures, emphasizing their evolution from domain-to-domain translation to the creation of original artistic content.

\subsection{GANs for Image Generation and Style Transfer}
Conditional GANs, introduced by Mirza et al., enable image generation conditioned on specific input variables, such as class labels \cite{mirza2014conditional}. This extension allows for controlled outputs but faces stability challenges, particularly with noisy or unbalanced data.

CycleGAN, developed by Zhu et al., facilitates unpaired image-to-image translation, enabling style transfer without requiring paired datasets \cite{zhu2017unpaired}. While effective, CycleGAN can distort image content during translation, affecting fidelity.
StyleGAN, proposed by Karras et al., generates high-resolution images with adjustable styles by disentangling style and content \cite{Karras_2020_CVPR}. Despite its success, StyleGAN demands significant computational resources and extensive training, limiting accessibility.

\subsection{Progressive GANs and Image Refinement}
Progressive Growing of GANs (ProGAN), introduced by Karras et al., improves training stability by starting with low-resolution images and progressively growing to higher resolutions \cite{karras2018progressivegrowinggansimproved}. This approach enhances detail in generated images but increases training time and resource demands as resolution grows.

Super Resolution GAN (SRGAN), proposed by Ledig et al., targets image super-resolution by generating high-resolution images from low-resolution inputs while maintaining perceptual quality \cite{8099502}. However, SRGAN can produce artifacts, particularly in complex textures.

\subsection{GAN Applications in Artistic Image Generation}
Generating artistic images from scratch remains challenging despite advances in style transfer. Neural style transfer, explored by Gatys et al., applies the artistic style of one image to another's content \cite{gatys2016image}, but relies on pre-existing content images.

ArtGAN, introduced by Tan et al., generates creative images by learning complex artistic features from training data \cite{tan2017artganartworksynthesisconditional}. It employs classification-aware loss functions to guide diverse artistic style generation but often struggles with inconsistent quality for certain styles and requires large datasets.

Creative Adversarial Networks (CANs), proposed by Elgammal et al., extend GANs to generate novel artwork by encouraging deviation from existing styles \cite{elgammal2017cancreativeadversarialnetworks}. While capable of originality, CANs can produce overly abstract outputs, requiring careful control.

Generating highly detailed and realistic artistic images faces persistent challenges, including instability, mode collapse, and balancing the generator and discriminator. Techniques like Wasserstein GANs (WGAN) \cite{arjovsky2017wasserstein} and Spectral Normalization \cite{miyato2018spectral} have been applied to stabilize training and enhance image diversity. This work incorporates these techniques to generate Monet-style images from scratch.

The complexity of fine art, characterized by detailed textures and intricate color compositions, adds to the challenges for GAN-based models. To address this, the proposed tiered GAN system uses multiple GANs in succession to progressively refine outputs, improving quality incrementally. Inspired by Amy Jang’s Monet CycleGAN tutorial \cite{jang2020cyclegan}, which demonstrates style transfer using unpaired datasets, this work generates Monet-style images directly from random noise rather than transforming existing images.

Building on the concept of progressive refinement introduced by ProGAN \cite{karras2018progressivegrowinggansimproved}, they proposed multi-stage GAN model that transforms random noise into detailed Monet-style images through collaborative refinement. Unlike image-to-image translation models such as CycleGAN; this approach focuses on generating original content from scratch.

Architectural choices were influenced by simpler datasets, such as Fashion MNIST \cite{renotte2022gan}, where GANs are effectively used to generate basic grayscale images. This work extends these principles to create detailed and stylistically accurate Monet-inspired artwork.

\section{Generative Adversarial Networks (GANs) Architecture}
GANs are a class of generative models consisting of two neural networks, namely \textit{Generator}  ($G$) and \textit{Discriminator} ($D$), trained simultaneously through adversarial learning. The primary objective of GANs is to generate synthetic data samples that are indistinguishable from real data samples.

\subsection{Generator ($G$)}
A Generator network, denoted as $G$, maps a noise vector $\mathbf{z} \in \mathbb{R}^d$ to a data sample $\mathbf{x}_{\text{fake}} \in \mathbb{R}^n$. The input noise vector $\mathbf{z}$ is typically sampled from a simple prior distribution such as a Gaussian or uniform distribution $p_{\mathbf{z}}(\mathbf{z})$. The mapping is mathematically expressed as:

\begin{equation}
    \mathbf{x}_{\text{fake}} = G(\mathbf{z}; \theta_G),
\end{equation}

where $\theta_G$ represents the learnable parameters of $G$. The goal of $G$ is to approximate the true data distribution $p_{\text{data}}(\mathbf{x})$ by generating realistic samples.

\subsection{Discriminator ($D$)}
A Discriminator network, denoted as $D$, acts as a binary classifier that distinguishes between real data samples $\mathbf{x}_{\text{real}}$ and generated (fake) data samples $\mathbf{x}_{\text{fake}}$. It outputs a scalar value $D(\mathbf{x}) \in [0, 1]$, representing the probability that the input data $\mathbf{x}$ is real. The function is defined as:

\begin{equation}
    D(\mathbf{x}; \theta_D),
\end{equation}

where $\theta_D$ represents the learnable parameters of $D$. The goal of $D$ is to maximize the probability of correctly identifying real data and minimize the probability of misclassifying generated data.

\subsection{Adversarial Objective}
The training process of GANs is structured as a two-player minimax game between $G$ and $D$. The following minimax objective function formalizes this:

\begin{equation}
    \begin{split}
    \min_G \max_D V(D, G) = & \, \mathbb{E}_{\mathbf{x} \sim p_{\text{data}}(\mathbf{x})} [\log D(\mathbf{x})] \\
    & + \mathbb{E}_{\mathbf{z} \sim p_{\mathbf{z}}(\mathbf{z})} [\log(1 - D(G(\mathbf{z})))].
    \end{split}
\end{equation}

here, the first term corresponds to $D$ maximizing the log probability of correctly identifying real data, while the second term corresponds to $D$ minimizing the probability of being deceived by fake samples.

\subsection{Loss Functions}
The loss functions for both networks are derived from the minimax objective.

\subsubsection{$D$ Loss}

The loss function for $D$ is defined as:

\begin{equation}
    \begin{split}
    L_D = - \mathbb{E}_{\mathbf{x} \sim p_{\text{data}}(\mathbf{x})} [\log D(\mathbf{x})] \\
    - \mathbb{E}_{\mathbf{z} \sim p_{\mathbf{z}}(\mathbf{z})} [\log(1 - D(G(\mathbf{z})))]
    \end{split}
\end{equation}

This encourages $D$ to output higher probabilities for real samples and lower probabilities for generated samples.

\subsubsection{$G$ Loss}
The loss function for $G$ is defined as:

\begin{equation}
    L_G = - \mathbb{E}_{\mathbf{z} \sim p_{\mathbf{z}}(\mathbf{z})} [\log D(G(\mathbf{z}))].
\end{equation}

This objective encourages $G$ to produce data that maximizes $D$'s output, effectively \textit{fooling} $D$ into classifying fake data as real.

\subsection{Training}
The training involves alternating updates to $D$ and $G$. Specifically, $D$ is updated to maximize its classification accuracy, while $G$ is updated to minimize $D$'s ability to distinguish real from fake samples. The training steps are as follows:

\begin{enumerate}
    \item Sample a batch of real data $\{\mathbf{x}_i\}_{i=1}^m$ from the data distribution $p_{\text{data}}(\mathbf{x})$.
    \item Sample a batch of noise vectors $\{\mathbf{z}_i\}_{i=1}^m$ from the noise distribution $p_{\mathbf{z}}(\mathbf{z})$.
    \item Compute $D$ loss $L_D$ and update $D$'s parameters $\theta_D$ using backpropagation.
    \item Compute $G$ loss $L_G$ and update $G$'s parameters $\theta_G$ using backpropagation.
\end{enumerate}

$D$'s objective is to maximize:

\begin{equation}
    \max_D \mathbb{E}_{\mathbf{x} \sim p_{\text{data}}(\mathbf{x})} [\log D(\mathbf{x})] + \mathbb{E}_{\mathbf{z} \sim p_{\mathbf{z}}(\mathbf{z})} [\log(1 - D(G(\mathbf{z}))],
\end{equation}

while $G$'s objective is to minimize:

\begin{equation}
    \min_G \mathbb{E}_{\mathbf{z} \sim p_{\mathbf{z}}(\mathbf{z})} [\log(1 - D(G(\mathbf{z})))],
\end{equation}

or equivalently:

\begin{equation}
    \min_G \mathbb{E}_{\mathbf{z} \sim p_{\mathbf{z}}(\mathbf{z})} [\log D(G(\mathbf{z}))].
\end{equation}

This formulation leads to an equilibrium where $G$'s distribution matches the real data distribution, making it challenging for $D$ to distinguish between real and fake samples.



\section{Methodology}
\subsection{Environment}
The software was developed using Python 3.9 and TensorFlow framework version 2.16.1. The experiment was conducted on Google Colab with 8 cores, utilizing an NVIDIA Tesla V100 GPU and 32 GB of RAM. The environment was supported by CUDA 11.8.0 and cuDNN version 8.8.0.

\subsection{Data Preprocessing}  
A dataset of RGB images with dimensions \( (256, 256, 3) \) was resized to \( (128, 128, 1) \) using nearest-neighbor interpolation. This resizing approach retained the recognizable features of the images, ensuring compatibility with our model's requirements at the chosen resolution. The dataset used, \textit{monet\_jpg}, consists of 300 Monet paintings sized \( 256 \times 256 \) in JPEG format and is publicly available on kaggle \cite{kaggle}.

\subsection{Model}
\begin{figure}[ht] 
    \centering
    \includegraphics[width=\columnwidth]{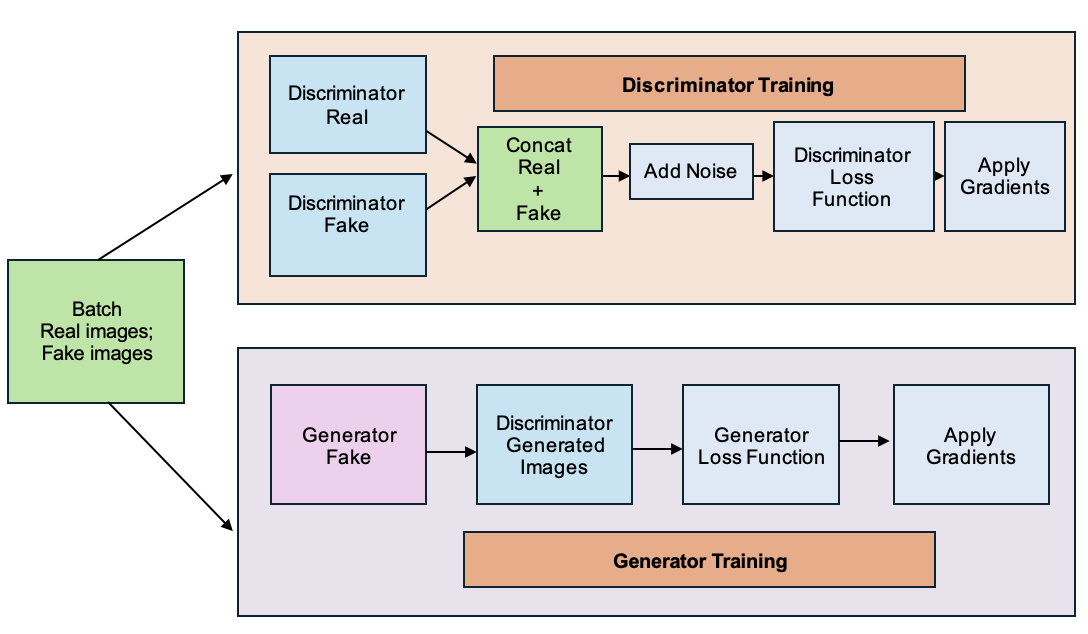}\vspace{-5pt}
    \caption{Different steps of \textit{Monet} GAN Model}\vspace{-15pt}
    \label{fig:Monet GAN Model}
\end{figure}

\begin{figure}[ht] 
    \centering
    \includegraphics[width=\columnwidth]{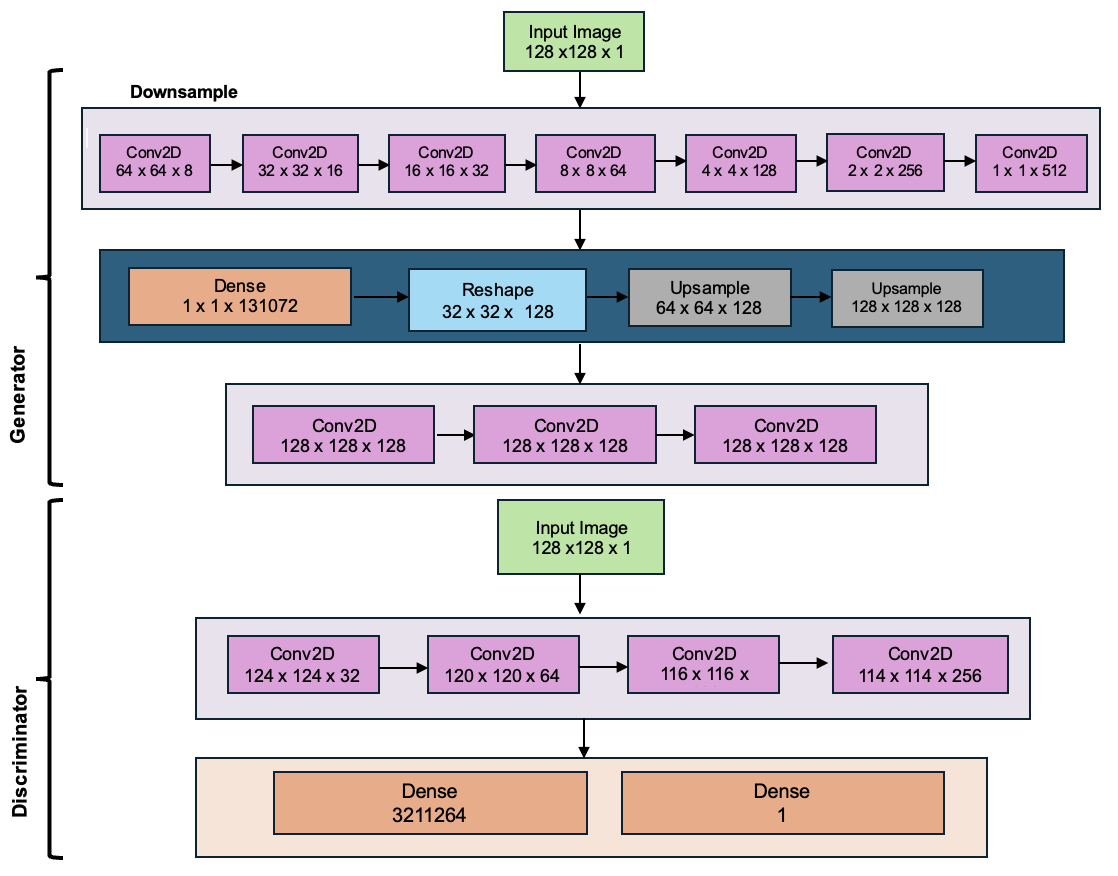} \vspace{-5pt}
    \caption{Generator and Discriminator Architecture}\vspace{-15pt}
    \label{fig:Monet Generator and Discriminator}
\end{figure}

The initial GAN model utilized random noise as input to generate full-size Monet paintings, establishing a foundation for further experimentation. Figure \ref{fig:Monet GAN Model} illustrates the Monet GAN architecture, featuring both $G$ and $D$, as shown in Figure \ref{fig:Monet Generator and Discriminator}. These are sequential models incorporating convolutional and dense layers. Leaky ReLU is employed as the activation function in the hidden layers, while the output layer of $D$ uses sigmoid activation for binary classification.

\begin{figure*}
    \centering
    \includegraphics[width=1\linewidth]{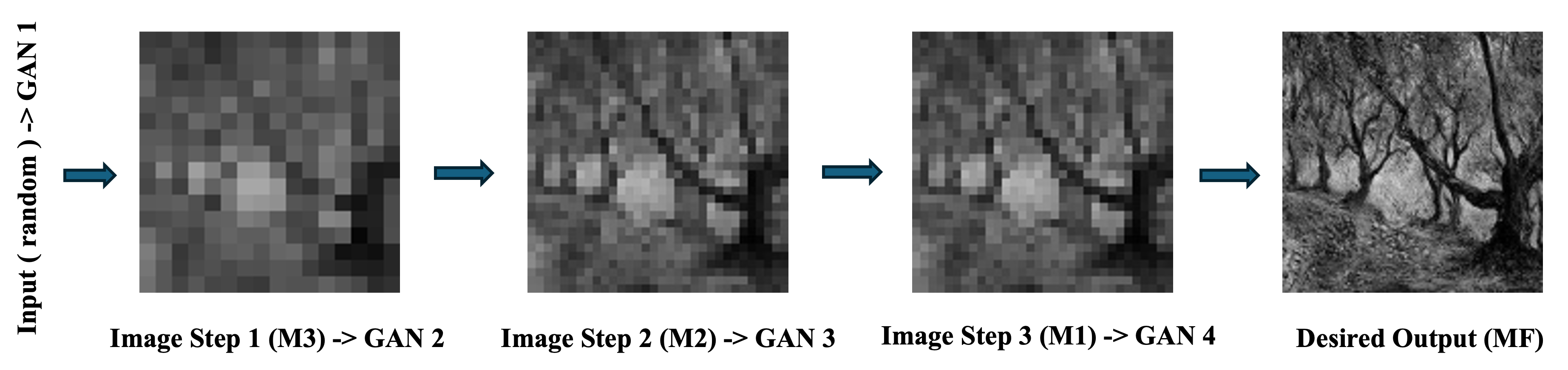}\vspace{-5pt}
    \caption{Progressive Refinement of Image Generation through Multiple Stages. $G$ begins with random noise (M3) and undergoes iterative enhancements across four GAN models, culminating in the final high-quality output (MF). Each stage represents a significant improvement in detail and quality, transitioning from M3 to M2, then to M1, before achieving the final image.}\vspace{-15pt}
    \label{fig:GAN}
\end{figure*}



\subsection{Input Strategy and Progressive Refinement}
The implementation explored various input strategies for $G$, beginning with random noise as input, which proved highly challenging. To address this, a multi-step refinement approach was developed, where $G$ begins with partially constructed images and progressively enhances them.

This method involved creating four datasets from the original dataset. These datasets, labeled MF, M1, M2, and M3, represent progressively lower-quality grayscale images with dimensions (128, 128, 1). MF contains the full grayscale image at half resolution, while M1, M2, and M3 are progressively downsampled versions with decreasing detail at each step.

$G$ refines the image through multiple stages, starting with random noise or a low-detail input:
\begin{itemize}
    \item \textbf{GAN 1}: Generates images for the M3 dataset (lowest quality).
    \item \textbf{GAN 2}: Refines M3 into M2.
    \item \textbf{GAN 3}: Refines M2 into M1.
    \item \textbf{GAN 4}: Produces the final image, MF.
\end{itemize}

The initial version of the model used random noise as the input, targeting the M3 dataset. A subsequent version incorporated non-random, downsampled images as input, enabling progressive refinement from M3 to M2, M2 to M1, and M1 to MF, as shown in Figure \ref{fig:GAN}.

\subsection{Training the Tiered GAN System}

The initial experiment aimed to generate Monet-style images from random noise using an input tensor of size \( (128, 128, 1) \). This experiment was conducted over 500 epochs, with a total training time of approximately 3 hours. Despite this effort, the results were unsatisfactory, as the generator (\( G \)) consistently produced black images. The limited number of training epochs, the relatively small dataset of 300 Monet paintings, and insufficient training time likely contributed to the model's poor performance, revealing the need for a more robust training approach.

To address these issues, a primary tiered GAN system was implemented, involving the sequential training of four separate GANs. Each GAN was trained for 2000 epochs, with the duration per epoch ranging from 22 to 27 seconds. Leveraging parallel processing on an NVIDIA Tesla V100 GPU with 8 CPU cores in the Google Colab environment, the model was able to efficiently perform computations such as convolutional operations and backpropagation. This parallelism reduced the overall training time, enabling the completion of 2000 epochs per GAN in approximately 12 hours, a significant improvement in computational efficiency. The testing phase, which involved generating images, took approximately 5 minutes per GAN, yielding outputs that progressively improved with each tier. This multi-stage strategy allowed the system to refine the generated outputs iteratively, enabling better learning from the dataset and overcoming the limitations of the initial experiment.

Parallel processing also facilitated higher scalability, enabling the system to handle more complex architectures and experiments without substantial increases in runtime. The tiered GAN approach, by leveraging optimized training strategies and sufficient resources, significantly enhanced the model's performance and quality of results.

To ensure balanced training between \( G \) and \( D \), both models were designed with relatively shallow architectures. The learning rate for \( D \) was set at one-tenth of \( G \)'s learning rate, specifically 0.0001 for \( G \) and 0.00001 for \( D \). The Adam optimizer was used for both models, with binary crossentropy as the loss function and a batch size of 8.

\begin{figure}[htbp] 
    \centering
    \begin{subfigure}[b]{0.45\columnwidth} 
        \centering
        \includegraphics[width=\linewidth]{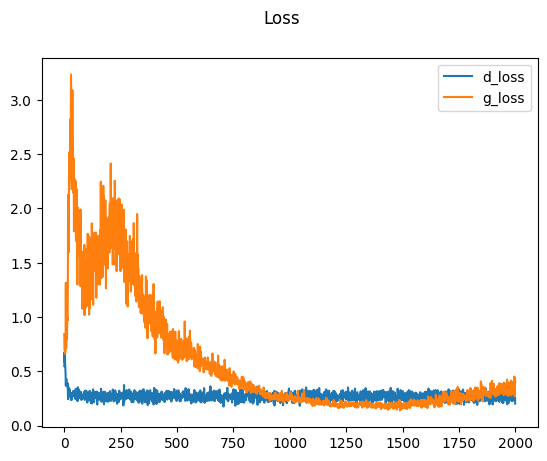}
        \caption{Random Noise to M3 Training Loss}
        \label{fig:image1}
    \end{subfigure}
    \hfill
    \begin{subfigure}[b]{0.45\columnwidth}
        \centering
        \includegraphics[width=\linewidth]{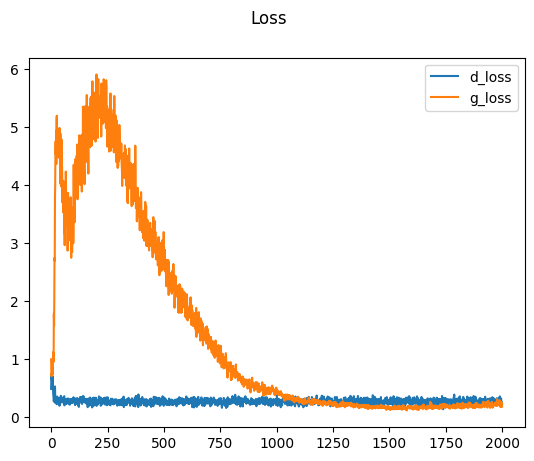}
        \caption{M3 to M2 Training Loss}
        \label{fig:image2}
    \end{subfigure}
    
    \vspace{0.3cm} 
    
    \begin{subfigure}[b]{0.45\columnwidth}
        \centering
        \includegraphics[width=\linewidth]{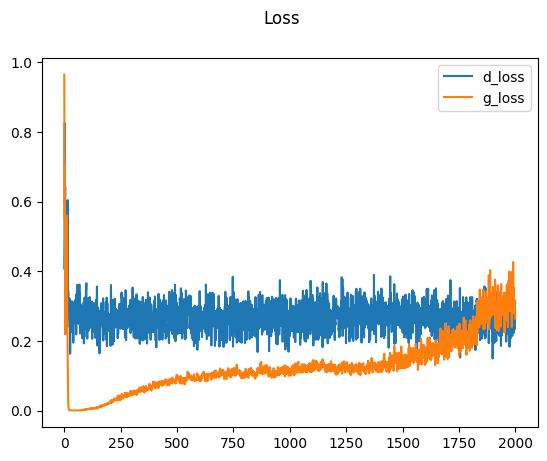}
        \caption{M2 to M1 Training Loss}
        \label{fig:image3}
    \end{subfigure}
    \hfill
    \begin{subfigure}[b]{0.45\columnwidth}
        \centering
        \includegraphics[width=\linewidth]{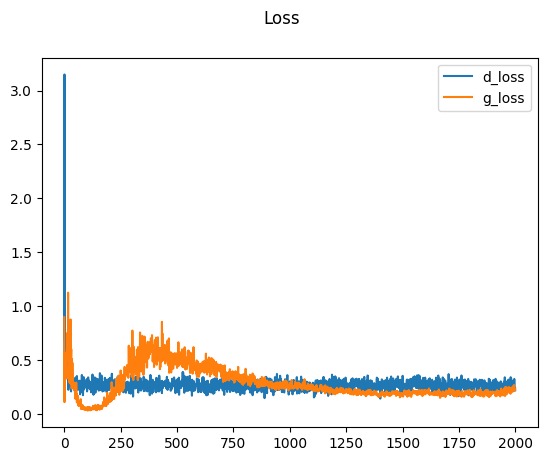}
        \caption{M1 to MF Training Loss}
        \label{fig:image4}
    \end{subfigure}
    
    \caption{Training loss curves for discriminator (\(d\_\text{loss}\)) and generator (\(g\_\text{loss}\)) across four phases(a-d): showing the adversarial dynamics and progressive model refinement.}

    \label{fig:images}
\end{figure}

\begin{figure}[ht]
    \centering
    \includegraphics[width=0.85\columnwidth]{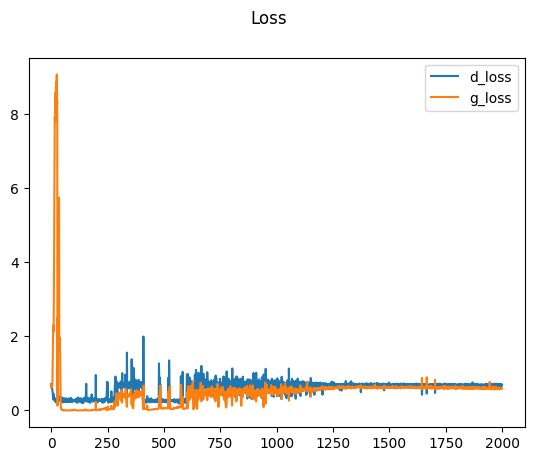}
    \caption{Training Loss for Small Input Vector with Upsampling}
    \label{fig:small_input_loss}
\end{figure}

\begin{figure*}[ht]
    \centering
    \includegraphics[width=1\textwidth]{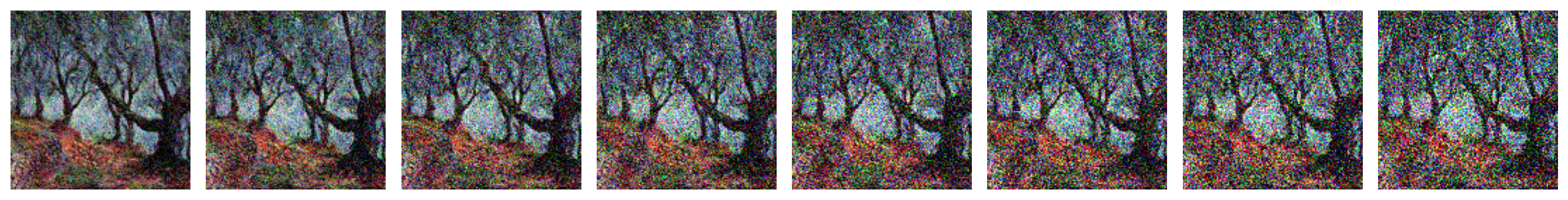}
    \caption{Generated Image from Small Input Vector with Upsampling}
    \label{fig:small_input_image}
\end{figure*}

\section{Results and discussion}

\subsection{Training Challenges}
During training, there were several challenges such as initial attempts to generate images from random noise yielded unsatisfactory results, indicating the need for extended training durations and a larger dataset. Memory constraints necessitated smaller input sizes, prompting adjustments to the generator architecture, including the use of convolutional layers for effective downsampling.

The tiered GAN system, designed to progressively refine images, initially failed to produce the desired outputs. Instead of Monet-like paintings, the system generated tensors with constant values of 1, highlighting the generator's inability to learn meaningful features. This required a thorough reevaluation of the model's architecture and training strategy. Figures \ref{fig:images} show the training loss for each GAN tier, with a high initial loss, particularly for the generator, which gradually decreases as both the generator and discriminator learn to balance each other. Specifically, Figure \ref{fig:image3} highlights the continued refinement of the model during this phase, where the generator and discriminator show improved synchronization, reducing loss and producing more coherent outputs.


\subsection{Model Re-evaluation and Alternative Approaches}

Issues were identified in the initial tiered GAN models, particularly with the downsampling convolution layers in the generator. To address these challenges, two alternative approaches were explored: first, using a small array of 128 random values and upsampling them to the target image size, as suggested by \cite{renotte2022gan}; second, replacing the input dense layers with convolutional layers to reduce memory usage and support larger input sizes.

Initial attempts with standard Conv2D layers proved ineffective. Drawing on methods from \cite{jang2020cyclegan} and \cite{karras2018progressive}, the approach was adjusted to use deconvolution layers (Conv2DTranspose) for upsampling, along with batch normalization to improve training stability and reduce convergence time. Despite these modifications, the generated images lacked clarity and artistic resemblance to Monet’s style.

The use of a smaller 128-element input array with upsampling showed better results. While clarity remained a limitation, the generated images exhibited recognizable patterns of light and dark areas, indicating that the model had started learning relevant features from the dataset. Figures \ref{fig:small_input_loss} and \ref{fig:small_input_image} show the training loss and an example of the generated images for this approach.

The generated images were evaluated for their resemblance to Monet’s style and ability to capture artistic details. Quantitative metrics like training loss monitored model performance, while human evaluation ensured alignment with Monet’s unique style. This combined approach of automated metrics and human assessment identified areas for further refinement in the model's architecture and training.

\section{Conclusion and Future Work}
This research introduced a tiered GAN architecture to employ multiple GANs sequentially for enhancing image quality, transforming low-quality images into refined representations of Monet's style. The training methodology efficiently handled large images using downsampling and convolutional layers, enabling high-quality artistic generation with limited computational resources.

Experimental results were mixed; while the system showed potential, it struggled to fully capture Monet's intricacies. The limited dataset of 300 images likely constrained the model's ability to learn complex artistic features. A larger and more diverse dataset could improve the model's performance and learning capability.

Future work will address these challenges through three strategies: (1) using larger datasets, augmented with bootstrap aggregating (bagging), to enhance prediction stability and robustness \cite{darweesh2018real}; (2) employing distributed computing inspired by Firebase for efficient processing and synchronization of large datasets \cite{eltehewy2023efficient}; and (3) incorporating pre-trained models via Transfer Learning to accelerate convergence and better capture Monet's artistic style.

These approaches aim to refine the tiered GAN system by leveraging real-time data handling and distributed computing, addressing current limitations in computational resources and dataset size.

\bibliographystyle{IEEEtran}
\bibliography{main.bib}

\begin{thebibliography}{10}
\providecommand{\url}[1]{#1}
\csname url@samestyle\endcsname
\providecommand{\newblock}{\relax}
\providecommand{\bibinfo}[2]{#2}
\providecommand{\BIBentrySTDinterwordspacing}{\spaceskip=0pt\relax}
\providecommand{\BIBentryALTinterwordstretchfactor}{4}
\providecommand{\BIBentryALTinterwordspacing}{\spaceskip=\fontdimen2\font plus
\BIBentryALTinterwordstretchfactor\fontdimen3\font minus \fontdimen4\font\relax}
\providecommand{\BIBforeignlanguage}[2]{{%
\expandafter\ifx\csname l@#1\endcsname\relax
\typeout{** WARNING: IEEEtran.bst: No hyphenation pattern has been}%
\typeout{** loaded for the language `#1'. Using the pattern for}%
\typeout{** the default language instead.}%
\else
\language=\csname l@#1\endcsname
\fi
#2}}
\providecommand{\BIBdecl}{\relax}
\BIBdecl

\bibitem{goodfellow2020generative}
I.~Goodfellow, J.~Pouget-Abadie, M.~Mirza, B.~Xu, D.~Warde-Farley, S.~Ozair, A.~Courville, and Y.~Bengio, ``Generative adversarial networks,'' \emph{Communications of the ACM}, vol.~63, no.~11, pp. 139--144, 2020.

\bibitem{hu2024tackling}
C.~Hu, T.~Tu, Y.~Gong, J.~Jiang, Z.~Zheng, and D.~Cheng, ``Tackling multiplayer interaction for federated generative adversarial networks,'' \emph{IEEE Transactions on Mobile Computing}, 2024.

\bibitem{chang2024enhancing}
Y.~Chang, ``Enhancing super resolution of oil painting patterns through optimization of unet architecture model,'' \emph{Soft Computing}, vol.~28, no.~2, pp. 1295--1316, 2024.

\bibitem{cai2021generative}
Z.~Cai, Z.~Xiong, H.~Xu, P.~Wang, W.~Li, and Y.~Pan, ``Generative adversarial networks: A survey toward private and secure applications,'' \emph{ACM Computing Surveys (CSUR)}, vol.~54, no.~6, pp. 1--38, 2021.

\bibitem{sharma2024generative}
P.~Sharma, M.~Kumar, H.~K. Sharma, and S.~M. Biju, ``Generative adversarial networks (gans): Introduction, taxonomy, variants, limitations, and applications,'' \emph{Multimedia Tools and Applications}, pp. 1--48, 2024.

\bibitem{bhati2024survey}
D.~Bhati, F.~Neha, and M.~Amiruzzaman, ``A survey on explainable artificial intelligence ({XAI}) techniques for visualizing deep learning models in medical imaging,'' \emph{Journal of Imaging}, vol.~10, no.~10, p. 239, 2024.

\bibitem{shi2024relu}
N.~Shi, Z.~Chen, L.~Chen, and R.~S. Lee, ``Relu-oscillator: Chaotic vgg10 model for real-time neural style transfer on painting authentication,'' \emph{Expert Systems with Applications}, p. 124510, 2024.

\bibitem{jang2020painter}
A.~Jang, A.~S. Uzsoy, and P.~Culliton, ``I’m something of a painter myself,'' \url{https://www.kaggle.com/competitions/gan-getting-started}, 2020.

\bibitem{vela2023improving}
L.~Vela, F.~Fuentes-Hurtado, and A.~Colomer, ``Improving the quality of image generation in art with top-k training and cyclic generative methods,'' \emph{Scientific Reports}, vol.~13, no.~1, p. 17764, 2023.

\bibitem{jin2022retracted}
X.~Jin, ``[retracted] art style transfer of oil painting based on parallel convolutional neural network,'' \emph{Security and Communication Networks}, vol. 2022, no.~1, p. 5087129, 2022.

\bibitem{mirza2014conditional}
M.~Mirza, ``Conditional generative adversarial nets,'' \emph{arXiv preprint arXiv:1411.1784}, 2014.

\bibitem{zhu2017unpaired}
J.-Y. Zhu, T.~Park, P.~Isola, and A.~A. Efros, ``Unpaired image-to-image translation using cycle-consistent adversarial networks,'' in \emph{Proceedings of the IEEE international conference on computer vision}, 2017, pp. 2223--2232.

\bibitem{Karras_2020_CVPR}
T.~Karras, S.~Laine, M.~Aittala, J.~Hellsten, J.~Lehtinen, and T.~Aila, ``Analyzing and improving the image quality of stylegan,'' in \emph{Proceedings of the IEEE/CVF Conference on Computer Vision and Pattern Recognition (CVPR)}, June 2020.

\bibitem{karras2018progressivegrowinggansimproved}
\BIBentryALTinterwordspacing
T.~Karras, T.~Aila, S.~Laine, and J.~Lehtinen, ``Progressive growing of gans for improved quality, stability, and variation,'' 2018. [Online]. Available: \url{https://arxiv.org/abs/1710.10196}
\BIBentrySTDinterwordspacing

\bibitem{8099502}
C.~Ledig, L.~Theis, F.~Huszár, J.~Caballero, A.~Cunningham, A.~Acosta, A.~Aitken, A.~Tejani, J.~Totz, Z.~Wang, and W.~Shi, ``Photo-realistic single image super-resolution using a generative adversarial network,'' in \emph{2017 IEEE Conference on Computer Vision and Pattern Recognition (CVPR)}, 2017, pp. 105--114.

\bibitem{gatys2016image}
L.~A. Gatys, A.~S. Ecker, and M.~Bethge, ``Image style transfer using convolutional neural networks,'' \emph{Proceedings of the IEEE conference on computer vision and pattern recognition}, pp. 2414--2423, 2016.

\bibitem{tan2017artganartworksynthesisconditional}
\BIBentryALTinterwordspacing
W.~R. Tan, C.~S. Chan, H.~Aguirre, and K.~Tanaka, ``Artgan: Artwork synthesis with conditional categorical gans,'' 2017. [Online]. Available: \url{https://arxiv.org/abs/1702.03410}
\BIBentrySTDinterwordspacing

\bibitem{elgammal2017cancreativeadversarialnetworks}
\BIBentryALTinterwordspacing
A.~Elgammal, B.~Liu, M.~Elhoseiny, and M.~Mazzone, ``Can: Creative adversarial networks, generating "art" by learning about styles and deviating from style norms,'' 2017. [Online]. Available: \url{https://arxiv.org/abs/1706.07068}
\BIBentrySTDinterwordspacing

\bibitem{arjovsky2017wasserstein}
M.~Arjovsky, S.~Chintala, and L.~Bottou, ``Wasserstein gan,'' \emph{arXiv preprint arXiv:1701.07875}, 2017.

\bibitem{miyato2018spectral}
T.~Miyato, T.~Kataoka, M.~Koyama, and Y.~Yoshida, ``Spectral normalization for generative adversarial networks,'' in \emph{International Conference on Learning Representations}, 2018.

\bibitem{jang2020cyclegan}
A.~Jang, ``Monet cyclegan tutorial,'' \url{https://www.kaggle.com/code/amyjang/monet-cyclegan-tutorial/notebook}, 2020, accessed 23 July 2023.

\bibitem{renotte2022gan}
\BIBentryALTinterwordspacing
N.~Renotte, ``Build a generative adversarial neural network with tensorflow and python | deep learning projects,'' YouTube, June 2022, accessed 18 July 2023. [Online]. Available: \url{https://www.youtube.com/watch?v=AALBGpLbj6Q}
\BIBentrySTDinterwordspacing

\bibitem{kaggle}
PodcastPrereamea, ``Cyclegan monet paintings,'' \url{https://www.kaggle.com/code/podcastprereamea/cyclegan-monet-paintings/data}, 2024.

\bibitem{karras2018progressive}
T.~Karras, T.~Aila, S.~Laine, and J.~Lehtinen, ``Progressive growing of gans for improved quality, stability, and variation,'' in \emph{International Conference on Learning Representations}, 2018.

\bibitem{darweesh2018real}
A.~Darweesh, A.~Abouelfarag, and R.~Kadry, ``Real time adaptive approach for image processing using mobile nodes,'' in \emph{2018 6th International Conference on Future Internet of Things and Cloud Workshops (FiCloudW)}.\hskip 1em plus 0.5em minus 0.4em\relax IEEE, 2018, pp. 158--163.

\bibitem{eltehewy2023efficient}
R.~Eltehewy, A.~Abouelfarag, and S.~N. Saleh, ``Efficient classification of imbalanced natural disasters data using generative adversarial networks for data augmentation,'' \emph{ISPRS International Journal of Geo-Information}, vol.~12, no.~6, p. 245, 2023.

\end{thebibliography}

\end{document}